\documentclass[letterpaper, 10 pt, conference]{ieeeconf}  

\IEEEoverridecommandlockouts                              

\usepackage{amsmath}
\usepackage{amssymb}
\usepackage{graphicx}
\usepackage{comment}
\usepackage{color}
\usepackage{cite}

\usepackage{booktabs,ragged2e}
\usepackage[flushleft]{threeparttable}
\usepackage[font=small,skip=5pt]{caption}
\usepackage[ruled,vlined]{algorithm2e}
\title{\LARGE \bf Pay Attention to How You Drive: Safe and Adaptive Model-Based Reinforcement Learning for Off-Road Driving}

\author{Sean J. Wang, Honghao Zhu, and Aaron M. Johnson
\thanks{All authors are with the Department of Mechanical Engineering, Carnegie Mellon University, Pittsburgh, PA, USA.
{\tt\small sjw2@andrew.cmu.edu, honghaoz@andrew.cmu.edu, amj2@andrew.cmu.edu}}}

\begin{document}
\maketitle

\begin{abstract}
Autonomous off-road driving is challenging as risky actions taken by the robot may lead to catastrophic damage. As such, developing controllers in simulation is often desirable as it provides a safer and more economical alternative. However, accurately modeling robot dynamics is difficult due to the complex robot dynamics and terrain interactions in unstructured environments. Domain randomization addresses this problem by randomizing simulation dynamics parameters, however this approach sacrifices performance for robustness leading to policies that are sub-optimal for any target dynamics. We introduce a novel model-based reinforcement learning approach that aims to balance robustness with adaptability. Our approach trains a System Identification Transformer (SIT) and an Adaptive Dynamics Model (ADM) under a variety of simulated dynamics. The SIT uses attention mechanisms to distill state-transition observations from the target system into a context vector, which provides an abstraction for its target dynamics. Conditioned on this, the ADM probabilistically models the system's dynamics. Online, we use a Risk-Aware Model Predictive Path Integral controller (MPPI) to safely control the robot under its current understanding of the dynamics. We demonstrate in simulation as well as in multiple real-world environments that this approach enables safer behaviors upon initialization and becomes less conservative (i.e.\ faster) as its understanding of the target system dynamics improves with more observations. In particular, our approach results in an approximately 41\% improvement in lap-time over the non-adaptive baseline while remaining safe across different environments.  
\end{abstract}

\section{Introduction}
Autonomous off-road driving has the potential to revolutionize applications such as environmental monitoring, planetary exploration, and agricultural automation by enabling robots to reach remote and challenging terrains \cite{6161683,4650755,bares1989ambler,bechar2016agricultural}. However, developing autonomous controllers for off-road driving can be challenging due to the dangerous nature of driving over uneven, unpredictable, and unstructured terrains. Inappropriate or misjudged actions can cause substantial damage to the robot, requiring expensive and time-intensive recovery and repair efforts.

Consequently, simulation has become instrumental in the development and validation of off-road driving algorithms. Beyond offering a risk-free environment for testing, simulations can operate faster than real-time, benefit from parallelization, and conduct trials autonomously. Simulation has been especially crucial in the development of model-free reinforcement learning algorithms \cite{zhang2018robot,9149720,9873725}, which aim to directly optimize a policy over many trials.

\begin{figure}[t] 
\centering 
\includegraphics[width=\linewidth]{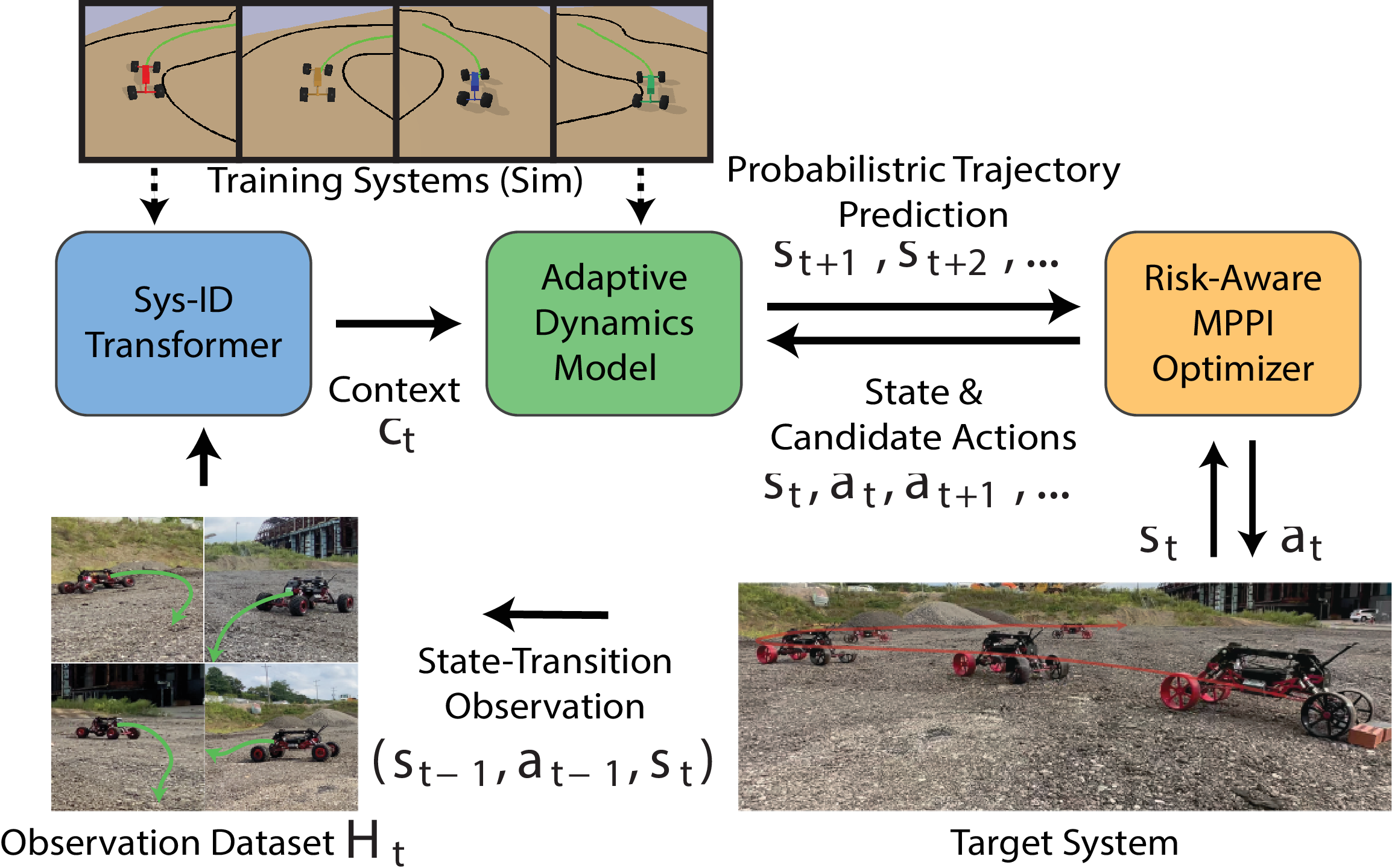}
\caption{Method Overview: The System Identification Transformer (SIT) and Adaptive Dynamics Model (ADM) are trained with randomized simulation dynamics to gain a probabilistic understanding of any target system's dynamics. The SIT leverages an attention mechanism to condense state-transition observations from the target system into a compact context vector. The ADM predicts state transition distributions conditioned on robot state, action, and context vector. Online, Risk Aware MPPI chooses safe actions according to the ADM's probabilistic predictions.}
\end{figure}

However, the performance of policies trained and validated in simulation do not always transfer to the real world. This discrepancy arises from the ``reality gap'' -- the inevitable differences between the simulated environment and the real world. Addressing these challenge requires effectively translating simulation-trained policies into the real-world, known as the ``sim2real'' transfer problem. While some methods aim to minimize the reality gap \cite{yu2017preparing, zhu2017model,tan2018sim, tsai2021droid,kaspar2020sim2real}, accurately modeling intricate dynamics of a robot interacting with a diverse range of unstructured terrains remains challenging. Robot dynamics are not only affected by the robot properties, such as weight distribution, tire friction coefficient, and motor models, but they are also affected by the unknown terrain properties including soil cohesion, dampness, or presence of debris.

Some approaches aim to train a policy that is effective on a wide range of dynamics, ideally including the dynamics of the real world system. In domain randomization \cite{tobin2017domain,tan2018sim,xie2021dynamics,peng2018sim}, simulation parameters are randomized during policy training to make the policy robust against variations in system dynamics. However, this robustness comes at the expense of conservative performance as the policy is not specifically tailored towards any particular system but generalized to all possible systems.

Alternatively, some approaches train a latent vector conditioned policy that can be adapted to some particular dynamics simply by identifying a suitable latent vector. In \cite{yu2020learning, yu2018policy, yu2019sim}, suitable latent vectors were found through optimization techniques such as CMA-ES \cite{hansen1995adaptation}. Although these approaches can tailor the policy towards the particular system, they still require trial and error to refine the policy, and may be unsafe while the policy is being refined. In \cite{kumar2021rma,yu2017preparing}, an auxiliary neural network is used to rapidly identify a suitable latent vector given a short fixed-length horizon of prior states and actions. While this allows for faster adaptation, the fixed-horizon input only utilizes recent observations for latent vector inference.
Furthermore, the model-free nature of these methods prohibits any interpretability with respect to the adaptation process or the resultant policy.

We propose a novel framework for sim2real transfer that balances robustness with adaptability. Our method follows the model-based reinforcement learning (MBRL) paradigm, where a probabilistic predictive dynamics model is first trained then used for decision making. We train the model in simulation with varying simulation parameters to make our model robust across a variety of system dynamics. Similar to prior methods \cite{kumar2021rma,yu2017preparing}, we train a neural network to extract a latent context vector to help adapt the policy to the target system's particular dynamics. In our approach, this neural network, called the System Identification Transformer (SIT), uses attention mechanisms to distill state-transition observations from the particular system into a context vector understanding of its particular dynamics. Unlike other approaches that condition a policy on this context vector, our approach instead conditions a dynamics model on the context vector. Given this context vector, the Adaptive Dynamics Model (ADM) probabilistically models the system's dynamics, capturing uncertainty both from the system's inherent stochasticity and from ambiguities due from insufficient state-transition observations. Online, we use a Risk-Aware Model Predictive Path Integral (RA-MPPI) controller \cite{yin2023risk} to safely control the robot under its current understanding of dynamics.

The remainder of this paper aims to validate the following hypotheses:
\begin{enumerate}
    \item Our proposed approach enables safer control in terms of the number of constraint violations, even when there is insufficient historical observation data (e.g.\ upon initialization).
    \item Leveraging the attention mechanism to extract context allows for continual improvement of the adapted policy (i.e.\ better lap times) as the number of state-transition observations increases.
    \item Using a risk aware MPPI controller reduces the number of constraint violations compared to a risk unaware controller with the same SIT and ADM models. 
\end{enumerate}

\section{Probabilistic Predictive Dynamics Model}
\label{sec:overview}
We formulate the autonomous off-road driving problem as a distribution of Markov decision processes (MDPs), where each real world environment is represented by a single MDP. For a given environment $i$, the problem is defined as $(\mathcal{S},\mathcal{A},\mathcal{P}_i,\mathcal{C}_i)$, where $\mathcal{S}$ is the state space, $\mathcal{A}$ is the action space, $\mathcal{P}_i(s_{t+1}|s_t,a_t)$ is the stochastic discrete-time transition dynamics from $s_t \in \mathcal{S}$ to $s_{t+1} \in \mathcal{S}$ under action $a_t \in \mathcal{A}$, and $\mathcal{C}_i(s,a)$ is the cost function for a given state-action pair. In this formulation, the state and action space is shared between environments, but the transition dynamics and cost function are unique to each environment. The function $\mathcal{P}_i$ is a member of the function space $\mathcal{F}$ that comprises all possible stochastic transition functions. We define $\mathcal{W}$ as the distribution over of the function space $\mathcal{F}$ which encompasses all potential dynamics functions the robot might encounter in the real world.

Note that it is impossible to perfectly simulate the unknown dynamics $\mathcal{P}_i$ for any given real world environment $i$, let alone the distribution $\mathcal{W}$ of all real world dynamics. We instead define a proxy distribution of dynamics in simulation, $\hat{\mathcal{W}}$, such that $\text{supp}(\mathcal{W}) \subseteq \text{supp}(\hat{\mathcal{W}})$. That is, all of the true dynamics in $\mathcal{W}$ lie within the range of dynamics functions represented in $\hat{\mathcal{W}}$. Using many cheap simulations sampled from $\hat{\mathcal{W}}$, we train a policy that can safely adapt to the particular system dynamics within $\text{supp}(\hat{\mathcal{W}})$, which includes all real world systems lying in $\mathcal{W}$.

Following the model-based reinforcement learning paradigm, we train a predictive model to approximate the probabilistic transition dynamics of any given system. The predictive model consists of two key components: the System Identification Transformer (SIT) and the Adaptive Dynamics Model (ADM). 
This model is then utilized for decision making, specifically using MPPI with a Conditional Value-at-Risk cost to drive the robot safely given the stochastic predictions from the predictive model.

\paragraph*{System Identification Transformer (SIT)}
The SIT, denoted by $\mathcal{T}_\theta$, identifies the dynamics of a given target system by analyzing prior state-transition observations collected on that system, denoted by $\mathcal{H}$, and extracting relevant information about the target system's dynamics into a latent context vector, denoted as $c$,
    \begin{equation}
        c_t = \mathcal{T}_\theta(\mathcal{H}_t)
    \end{equation}
In this formulation, state-transition observations collected for a target system at time $t$ are,
    \begin{equation}
        \mathcal{H}_t = \{(s_i,a_i,s_{i+1})|i<t-1\},
    \end{equation}

We use a transformer network \cite{vaswani2017attention} for the SIT due to several advantages it offers. Transformers can natively accommodate sequences of varying lengths by utilizing self-attention mechanisms to selectively focus on specific segments of the input sequence. These advantages are crucial for our application since state-transition observation sequences expand with the system's run-time. Furthermore, not all state-transition observations are of equal significance (e.g., periods when the robot remains stationary may offer minimal insights), so this selective focus ensures the extracted context is most representative of the system's dynamics.

The SIT's architecture mirrors the encoder component from \cite{vaswani2017attention}. It comprises of a series of identical layers, each featuring a multi-head self-attention sub-layer followed by a position-wise, fully connected feed-forward network sub-layer. Each sub-layer incorporates a residual connection \cite{he2016deep} followed by a layer normalization \cite{ba2016layer}. Unlike the original design, we opted not to use positional encoding for the input sequences. In our application, the order of (state, action, state-transition) observations is unimportant, and incorporating positional encoding negatively impacted performance. Finally, we aggregated the vector outputs from the last layer by taking their mean, resulting in a single context vector. This compact representation, $c\in \mathbb{R}^{32}$ for our implementation, encapsulates the essence of all prior state-transition observations.

\paragraph*{Adaptive Dynamics Model (ADM)}
The ADM provides a probabilistic understanding of the robot's dynamics based on the context vector extracted by the SIT. The ADM, denoted as $\mathcal{P}_\theta(s_{t+1}|s_t,a_t,c_t)$, is trained to predict state-transition distributions conditioned on the robot's current state, action, and context vector $c_t$ extracted by the SIT. By predicting state-transitions as probability distributions, the ADM can capture uncertainty inherent to the non-deterministic system as well as ambiguities resulting from limited state-transition observations.

Similar to \cite{wang2021rough}, the adaptive dynamics model can be used to predict a trajectory distribution for the robot by sequentially iterating through each time step of the prediction horizon and chaining samples from the predicted state-transitions distribution, as is done in Algorithm \ref{alg:cvar}. We chose to use a Long Short Term Memory (LSTM) architecture \cite{hochreiter1997long}, which inherently captures temporal dependencies across state-transition sequences. For our implementation, the LSTM is followed by a fully connected network to predict a multivariate Gaussian state-transition distribution, parameterized by its mean and lower triangular terms of the LU decomposition of its covariance matrix.

\section{Risk-Aware Model Predictive Path Integral Control}
\label{sec:mppi}
In this section, we describe how controls can be made robust against the uncertainty in the probabilistic output of the SIT and ADM. This allows the robot to drive safely even when it is unsure about its dynamics while improving performance as its understanding improves with more state-transition observations.

\paragraph*{Track Driving Problem} For our application, the robot is tasked with driving down different tracks. 
Each track is defined by a $path$ (its center line) and a fixed width $w$. Given $path$, we structure the task as the following constrained optimization problem,
\begin{align}
\underset{a_{t_0},...,a_{t_f}}{\text{minimize}} \quad & L_{path}(s_{t_f+1}) \label{eq:objective}\\ 
\text{subject to:} \quad & s_{t+1} \sim \mathcal{P}_i(s_{t+1}|s_t,a_t) \label{eq:opt_dyn_true}\\
& D_{path}(s_t) \leq w \label{eq:track_constraint}\\
& |\Ddot{s}_{t,lateral}| \leq A \label{eq:lat_accel_constraint},
\end{align}
where $\mathcal{L}_{path}(s)$ denotes the distance of state $s$ along $path$, $D_{path}(s)$ denotes the distance of state $s$ from $path$, and $\Ddot{s}_{t,lateral}$ denotes the lateral component of the robot's acceleration (calculated through numerical differentiation). Intuitively, the robot's task is to make as much progress down the track \eqref{eq:objective}, while staying on track \eqref{eq:track_constraint}, and keeping lateral acceleration under a threshold to prevent it from rolling over \eqref{eq:lat_accel_constraint}, subject to the stochastic dynamics \eqref{eq:opt_dyn_true}.

\paragraph*{Robust Controls}

While numerous methods exist for robust control of systems with probabilistic dynamics, e.g.\ \cite{gandhi2021robust,chua2018deep}, we use Model Predictive Path Integral (MPPI) \cite{williams2018information} with a Conditional Value-at-Risk (CVaR) cost to avoid risky actions, similar to \cite{yin2023risk}. 

MPPI is a variant of Model Predictive Control (MPC) that relies on a sampling-based approach for trajectory optimization. During each MPPI optimization iteration, candidate action sequences are sampled from a distribution centered around the previous solution. The cost associated with each candidate action sequence is evaluated by simulating the system with a predictive model. The solution is then updated by weighting the candidate actions based on their costs.

To minimize constraint violation within MPPI, we use the relaxed logarithmic barrier function introduced in \cite{feller2016relaxed}. This function reformulates a constraint of the form $z \geq 0$, into the following as an additional cost term:
\begin{align}
    \hat{B}(z) &= \begin{cases}
      -ln(z) & z > \delta \\
      \beta_e(z;\delta) & z \leq \delta
    \end{cases}\\
    \beta_e(z;\delta) &= \exp{(1-\frac{z}{\delta})} - 1 - \ln{\delta}
\end{align}

In our approach, we enhance the robustness of MPPI against uncertainties in system dynamics by incorporating a CVaR cost, Algorithm~\ref{alg:cvar}. The CVaR cost quantifies the expected cost in the worst $\alpha$ percent of scenarios. To calculate the CVaR cost for each candidate action sequence, we perform multiple trajectory simulations using our stochastic dynamics model (ADM) and average the cost of the worst-performing trajectories. This enables the optimizer to be risk-aware when choosing actions.

\begin{algorithm}[t]
\caption{Calculating CVaR Cost}
\label{alg:cvar}
\SetKwInOut{Input}{Input}
\SetKwInOut{Output}{Output}
\Input{Initial State: $s_{t_0}$,\\
Context Vector: $c_{t_0} = \mathcal{T}_\theta(\mathcal{H}_{t_0})$\\
Candidate Actions: $a_{t_0}, a_{t_1}, \ldots, a_{t_f}$,\\
Number of Stochastic Evaluations: $N$\\
Confidence Level: $\alpha$}
\Output{CVaR cost}
\BlankLine
\For{$j \leftarrow 1$ \KwTo $N$}{
    $\hat{s}_{t_0} \gets s_{t_0}$\\
    $J_j \gets 0$\\
    \For{$t \leftarrow t_0$ \KwTo $t_f$}{
        $\hat{s}_{t+1} \sim \mathcal{P}_\theta(\hat{s}_{t+1}| \hat{s}_t, a_t, c_t)$\\
        $J_j \gets J_j + \mathcal{C}(s_t, a_t)$
    }
}
\BlankLine
\Return{Average of top $\lceil \alpha \cdot N \rceil$ values of $ J$}
\end{algorithm}

\section{Training in Simulation}
We train the SIT and ADM solely in simulation. However, instead of using one simulated system, we sample a large number of simulated systems from the distribution $\hat{\mathcal{W}}$, created by randomly varying physical parameters in simulation. By doing so, we train the SIT and ADM to adapt to a wide variety of systems including real world systems from the distribution $\mathcal{W}$. During training, we cycle between a data collection phase and a model training phase.

\paragraph*{Data Collection}
During the data collection phase, we first generate a set of new systems in simulation using PyBullet. To generate a new system, we randomize link dimensions, link inertial terms, scaling of steering and throttle commands, motor torque and PID values, contact parameters (friction, stiffness, and damping), and suspension parameters (limits, stiffness, damping). For each system, we collect a set of trajectories by driving the system using the current SIT and ADM models within the Risk Aware MPPI framework. At each time step during driving, the policy is adapted to the particular system by feeding all prior collect data on that system into the SIT.

\paragraph*{Neural Network Training}
During the model training phase, we sample a system and time step from the dataset and use the SIT and ADM to predict the state-transition given the robot's current state, action, and all state-transition observations collected on the particular system prior to that time step. We update the neural network parameters of SIT and ADM using a negative log-likelihood loss with an Adam optimizer \cite{kingma2014adam}.

\section{Experimental Results}
We compare our method against a different baselines in simulation and on a real world robot. In simulation, we run large statistical tests comparing the performance metrics of different approaches on newly generated systems and tracks, none of which were seen during training. On the real world system, we evaluate whether the trained model and resultant policy can safely adapt to different real world systems. We vary the dynamics of the real world system by changing the robot's configuration and varying the type of terrain used. Both the simulated and real world systems use a four-wheeled robot with flexible solid-axle suspension and all wheel steering. For the real world system, MPPI controls was ran on board at 10 Hz using a NVIDIA GeForce RTX 2060 GPU.

\subsection{Fast and Continual Adaptation to New Dynamics}
\label{sec:exp_1}
In the first experiment, we evaluate this method's ability to generate a safe and effective policy for a new system upon initialization and then continually adjust that policy to better adapt to the target system. For each newly generated system, we run trials over randomly generated tracks starting with no state-transition observations. These new state-transition observations created by driving the system are collected and used to adapt the model at every time step.

For the baseline comparison, we use a model-based reinforcement learning policy where the neural network dynamics model is reinitialized for each new system and trained using only data collected on that particular system. In this baseline approach, the dynamics model uses the same architecture as our Adaptable Dynamics Model, but is given a fixed zero vector for the context input. We collect training data by driving the robot using the baseline model and retraining the model every 250 time steps.

We evaluated the performance of both methods as a function of time steps collected for training or adaptation. We fix the models created given different amounts of data and use them to drive the robot down a new test track. The test track is fixed between all methods and models for a particular system, but varied for the different systems or trials. Note that for our method, we allowed the model to continue adapting on the test track run since adaptation involved simple SIT inference, which could be computed at each time step.

\begin{figure}[t] 
\centering 
\includegraphics[width=\linewidth]{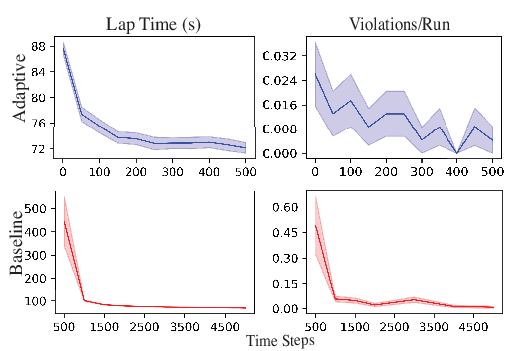}
\caption{Adaptive Method vs. Baseline Method. For the baseline, a new policy was trained on each target system. The standard error is shown with shaded region. For each time step at fixed intervals, we take the model trained at that time step to run each system on a test track in simulation. We averaged out the lap time and violations across all systems as it completed the track. } 
\label{fig:train_test} 
\end{figure}

During each evaluation, we record the lap time (in time steps of 0.1 seconds) needed to complete the test track as well as the number of constraint violations (either the robot driving off track or exceeding the lateral acceleration limit). We also record the number of times the robot made no progress, or was stationary for too long, due to MPPI struggling to find a non-trivial solution. In cases where the robot makes no progress or violates the constraints, the robot is reset to the center of the track at the last progress point and allowed to continue. The average lap time and number of constraint violations for both methods across 230 systems are shown in Fig.~\ref{fig:train_test}.

Compared to the baseline, our method was able to reach much higher levels of performance in low-data regimes. With our method, the robot was able to drive down the test track even when initialized with zero data. With the baseline method, we were unable to evaluate its performance with less than 500 time steps of training data, as the robot often could not finish the track. When comparing our method given zero data and the baseline given 500 time steps of data, our method had a much faster lap time and exhibited many fewer constraint violations. Furthermore, the baseline method averaged 3.4 incidents of no progress per test track run, where the robot needed to be reset due to not making any progress. In comparison, our method averaged $0.004$ incidents. Unlike our approach, the baseline approach is impractical to deploy on real world systems due to the high number of constraint violations and resets needed in low-data regimes. This evidence supports \textbf{hypothesis 1}, since our approach enables safer control in absence of historical observation data.

For both methods, the policy's performance improved with more training data. When given 5000 time steps of data, the baseline method exhibited an average lap time of $69.43$ time steps and averaged $0.0087$ constraint violations. In contrast, after only 500 time steps our approach had an average lap time of $72.17$ and averaged $0.0043$ constraint violations.

By using attention mechanisms in the SIT, our approach can use variable length state-transition observation sequences to tailor the ADM to a particular system. This allows for continual improvement of the policy for a potentially long period of time, where the observation sequence is long. This is shown in our experiment (Fig. \ref{fig:train_test}), where our method exhibits gradual performance improvements from 0 to 500 time steps of data. Furthermore, at 500 time steps of data, the performance of our method is comparable to the performance limits of training a policy from scratch for the particular system. This supports \textbf{hypothesis 2}, since the adaptive method continually improves in lap-time performance as the number of state-transition observations increases.

\subsection{Safety During Adaptation}
We evaluated our method's ability to remain safe during initial periods of adaptation by comparing it to a baseline that did not consider uncertainty in MPPI. During MPPI, this baseline calculated the cost of an action sequence by predicting the resulting trajectory using the deterministic transition model $\hat{s}_{t+1} = \mathbb{E}[\mathcal{P}_\theta(\hat{s}_{t+1}| \hat{s}_t, a_t, c_t)]$. This is in contrast to our method that calculates a CVaR cost based on predicting multiple possible trajectories under the stochastic dynamics $\mathcal{P}_\theta(s_{t+1}| s_t, a_t, c_t)$, described in Sec.~\ref{sec:mppi}.

\begin{figure}[t] 
\centering 
\includegraphics[width=\linewidth, keepaspectratio]{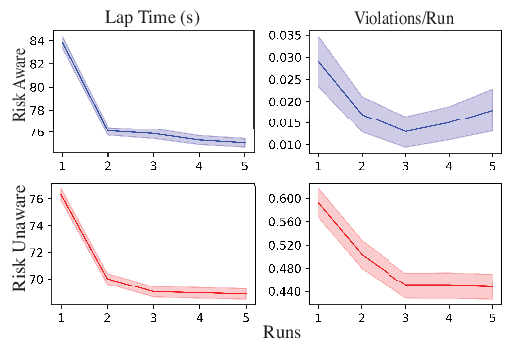} 
\caption{Risk Aware vs Risk Unaware MPPI with Adaptive Model. Lap time and number of violations per run are shown, with the average over all systems as the solid line and the standard error shown as the shaded region.} 
\label{fig:robust_test} 
\end{figure}

We compared the two methods by generating 1000 new systems in simulation and one track per system. For each system, we use both methods to drive the robot down the same track 5 times, starting with zero state-transition observations on the first run and adapting the model at each time step throughout the 5 runs. For the two methods, we plot the average lap time and number of constraint violation for the 5 runs, across the 1000 systems, in Fig.~\ref{fig:robust_test}.

Our method exhibited far fewer constraint violations than the baseline method that did not use risk aware MPPI. The average number of violations among all runs were $0.015$ for our method and $0.49$ for the risk unaware MPPI method. The risk unaware MPPI method exhibited more violations on the first run, with an average of $0.59$, than on the last run, with an average of $0.45$, due to the model adapting and improving. For both methods, the lap time dramatically improved from the first run to the second run, but had minimal improvements after the second run. We attribute this to the robot driving down the same track for all runs leading to saturation of useful information that could be extracted after the first run. The risk unaware MPPI method had a significantly faster lap time than the risk aware MPPI method. However, it achieved faster lap times by driving aggressively off track and leveraging the penalty-free resets to the middle of the track whenever a constraint was violated. This evidence supports \textbf{hypothesis 3}, since the risk aware MPPI is shown to have significantly fewer constraints compared to the risk unaware version.

\subsection{Sim2real Transfer}
\begin{figure}[t] 
\centering 
\includegraphics[width=\linewidth, keepaspectratio]{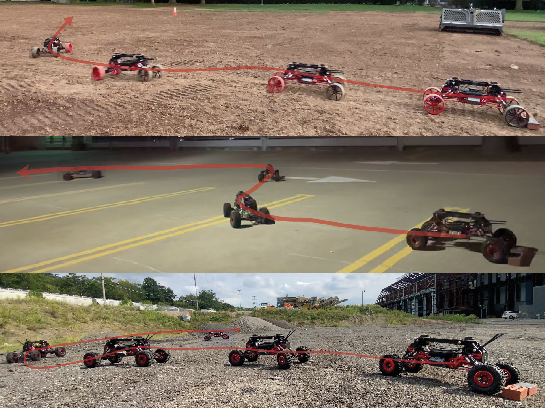} 
\caption{Sim2real Experiments. From top to bottom, the wheeled robot driving on dirt, concrete, and gravel. The red line indicates the predefined track for each trial. The tires were changed from compliant rubber tires to hard plastic tires in the dirt experiment shown.} 
\label{fig:realexp} 
\end{figure}

In this experiment, we evaluated the ability of our method to transfer to real world dynamics, Fig.~\ref{fig:realexp}. As a baseline, we trained a model-based reinforcement learning policy in simulation using only the fixed nominal dynamics (simulation parameters were not varied). The training procedure for this baseline method followed closely to that from Sec.~\ref{sec:exp_1}. We then ran the policy from both methods on a real world robot. Between trials, we introduced variations to the system's dynamics by changing the terrain type (concrete, dirt, and gravel) and the robot's configuration by changing the scaling of steering and throttle commands as well as swapping the standard rubber tires with low friction PLA 3D printed tires. For each new system dynamics, we reinitialized our method and allowed it to adapt to the new system's dynamics. The baseline method was fixed and therefore was not retrained whenever the system changed. In total, we ran 10 trials of each.

\begin{figure}[t] 
\centering 
\includegraphics[width=\linewidth, keepaspectratio]{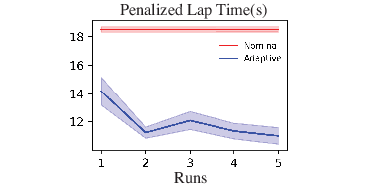} 
\caption{Adaptive Model vs. Nominal Model for Sim2real. For the adaptive model, we show average lap time of different systems across each run. Given the nominal model's inability to adapt leading to no difference in method between runs, we plot the average across all runs and systems. Shaded region indicates standard error.} 
\label{fig:sim2real} 
\end{figure}

For each trial, we used both methods to drive the robot down a fixed track 5 times. For our adaptive method, the robot was given no state-transition observations at the start of the first run, but allowed to adapt using collected observations at each time step throughout the 5 runs. For the baseline method, performing more runs had no effect since there was no mechanism for adaptation. As such, we averaged the performance over all runs for the baseline method. For all runs, the robot was automatically stopped anytime a constraint was violated and manually placed on the center of the track. For every reset, we assigned a 10 second penalty, as the manually resets usually took longer than 10 seconds. The penalized lap times for both methods are shown in Fig. \ref{fig:sim2real}.

Our method completed all runs with a 100\% success rate, where success is defined as completing the track with no constraint violations. This is much higher than the baseline, which had a 40\% success rate. Furthermore, we ran a paired t-test between the first and second run's laptime for our method and found significant improvement for the second run with a p value of 0.017. However, none of the successive runs showed any further significant improvement ($p < 0.05$) from the second run. Again, we attribute this to the fixed track leading to saturation of useful adaptation information after the first run. For the baseline method, which used a non adaptive model, there was no statistically significant difference in lap times between runs. This provides additional evidence for \textbf{hypotheses 1} and \textbf{2}, since the adaptive approach is shown to remain safe in low-data regimes, while continually improving as it collects more observation data across different real-world environments.

\section{Discussion \& Conclusion}
In this paper, we propose a novel sim2real transfer framework that balances robustness with adaptability. Our approach trains two neural network models, the System Identification Transformer (SIT) and the Adaptive Dynamics Model (ADM), in simulation while randomizing simulation dynamics parameters. The SIT leverages attention mechanisms to distill state-transition observations collected on the target system into a context vector, which succinctly encodes knowledge about the particular system's dynamics. The ADM predicts state-transition distributions given the robot's current state, action, and context vector from the SIT. Together, the SIT and ADM capture a probabilistic understanding of a target system's dynamics from state-transition observations on the target system. In real-time, our framework utilizes MPPI combined with a CVaR cost to safely control the system under its current understanding of dynamics.

Our approach ensures safe controls even with sparse observations by capturing a probabilistic understanding of the robot's dynamics, thereby enhancing control robustness. Furthermore, our method facilitates continual adaptation and performance enhancement as the robot operates and accumulates more state-transition observations. This adaptability stems from the attention mechanisms in the SIT, which can process variable-length observations and focus on pertinent segments of extended sequences to distill insights about the system's dynamics.

In our experiments, both in simulation and in the real world, we demonstrate the effectiveness of our approach in safely driving unseen systems right from the initialization, with zero state-transition observations. Moreover, as more state-transition observations were gathered, our method exhibited marked performance enhancements, indicating its adaptability to the dynamics of the target system. This adaptability was particularly evident in trials where observations were collected across different tracks. Each track appeared to enrich the system's understanding, subsequently elevating its performance.

One limitation of our approach is the presumption of static system dynamics during execution. However, in real-world settings, a robot's dynamics can often change due to transitions between different terrains, wear and tear of the hardware, and more. Future work could incorporate mechanisms to detect these dynamic shifts, and subsequently re-initializing the adaptation process. Additionally, refinements to the current SIT can lead to potential improvement of adaptation to such dynamic changes.

\addtolength{\textheight}{-1cm}   
                                  
\bibliographystyle{IEEEtran}
\bibliography{references}

\begin{thebibliography}{10}
\providecommand{\url}[1]{#1}
\csname url@samestyle\endcsname
\providecommand{\newblock}{\relax}
\providecommand{\bibinfo}[2]{#2}
\providecommand{\BIBentrySTDinterwordspacing}{\spaceskip=0pt\relax}
\providecommand{\BIBentryALTinterwordstretchfactor}{4}
\providecommand{\BIBentryALTinterwordspacing}{\spaceskip=\fontdimen2\font plus
\BIBentryALTinterwordstretchfactor\fontdimen3\font minus
  \fontdimen4\font\relax}
\providecommand{\BIBforeignlanguage}[2]{{%
\expandafter\ifx\csname l@#1\endcsname\relax
\typeout{** WARNING: IEEEtran.bst: No hyphenation pattern has been}%
\typeout{** loaded for the language `#1'. Using the pattern for}%
\typeout{** the default language instead.}%
\else
\language=\csname l@#1\endcsname
\fi
#2}}
\providecommand{\BIBdecl}{\relax}
\BIBdecl

\bibitem{6161683}
M.~Dunbabin and L.~Marques, ``Robots for environmental monitoring: Significant
  advancements and applications,'' \emph{IEEE Robotics \& Automation Magazine},
  vol.~19, no.~1, pp. 24--39, 2012.

\bibitem{4650755}
M.~Trincavelli, M.~Reggente, S.~Coradeschi, A.~Loutfi, H.~Ishida, and A.~J.
  Lilienthal, ``Towards environmental monitoring with mobile robots,'' in
  \emph{IEEE/RSJ International Conference on Intelligent Robots and Systems},
  2008, pp. 2210--2215.

\bibitem{bares1989ambler}
J.~Bares, M.~Hebert, T.~Kanade, E.~Krotkov, T.~Mitchell, R.~Simmons, and
  W.~Whittaker, ``Ambler: An autonomous rover for planetary exploration,''
  \emph{Computer}, vol.~22, no.~6, pp. 18--26, 1989.

\bibitem{bechar2016agricultural}
A.~Bechar and C.~Vigneault, ``Agricultural robots for field operations:
  Concepts and components,'' \emph{Biosystems Engineering}, vol. 149, pp.
  94--111, 2016.

\bibitem{zhang2018robot}
K.~Zhang, F.~Niroui, M.~Ficocelli, and G.~Nejat, ``Robot navigation of
  environments with unknown rough terrain using deep reinforcement learning,''
  in \emph{IEEE International Symposium on Safety, Security, and Rescue
  Robotics}, 2018, pp. 1--7.

\bibitem{9149720}
S.~Josef and A.~Degani, ``Deep reinforcement learning for safe local planning
  of a ground vehicle in unknown rough terrain,'' \emph{IEEE Robotics and
  Automation Letters}, vol.~5, no.~4, pp. 6748--6755, 2020.

\bibitem{9873725}
B.~Zhou, J.~Yi, and X.~Zhang, ``Learning to navigate on the rough terrain: A
  multi-modal deep reinforcement learning approach,'' in \emph{IEEE
  International Conference on Power, Intelligent Computing and Systems}, 2022,
  pp. 189--194.

\bibitem{yu2017preparing}
W.~Yu, J.~Tan, C.~K. Liu, and G.~Turk, ``Preparing for the unknown: Learning a
  universal policy with online system identification,'' in \emph{Robotics:
  Science and Systems}, 2017.

\bibitem{zhu2017model}
S.~Zhu, A.~Kimmel, K.~E. Bekris, and A.~Boularias, ``Fast model identification
  via physics engines for data-efficient policy search,'' in
  \emph{International Joint Conference on Artificial Intelligence}, 2018, pp.
  3249--3256.

\bibitem{tan2018sim}
J.~Tan, T.~Zhang, E.~Coumans, A.~Iscen, Y.~Bai, D.~Hafner, S.~Bohez, and
  V.~Vanhoucke, ``Sim-to-real: Learning agile locomotion for quadruped
  robots,'' in \emph{Robotics: Science and Systems}, 2018.

\bibitem{tsai2021droid}
Y.-Y. Tsai, H.~Xu, Z.~Ding, C.~Zhang, E.~Johns, and B.~Huang, ``Droid:
  Minimizing the reality gap using single-shot human demonstration,''
  \emph{IEEE Robotics and Automation Letters}, vol.~6, no.~2, pp. 3168--3175,
  2021.

\bibitem{kaspar2020sim2real}
M.~Kaspar, J.~D.~M. Osorio, and J.~Bock, ``Sim2real transfer for reinforcement
  learning without dynamics randomization,'' in \emph{IEEE/RSJ International
  Conference on Intelligent Robots and Systems}, 2020, pp. 4383--4388.

\bibitem{tobin2017domain}
J.~Tobin, R.~Fong, A.~Ray, J.~Schneider, W.~Zaremba, and P.~Abbeel, ``Domain
  randomization for transferring deep neural networks from simulation to the
  real world,'' in \emph{IEEE/RSJ International Conference on Intelligent
  Robots and Systems}, 2017, pp. 23--30.

\bibitem{xie2021dynamics}
Z.~Xie, X.~Da, M.~Van~de Panne, B.~Babich, and A.~Garg, ``Dynamics
  randomization revisited: A case study for quadrupedal locomotion,'' in
  \emph{IEEE International Conference on Robotics and Automation}, 2021, pp.
  4955--4961.

\bibitem{peng2018sim}
X.~B. Peng, M.~Andrychowicz, W.~Zaremba, and P.~Abbeel, ``Sim-to-real transfer
  of robotic control with dynamics randomization,'' in \emph{IEEE International
  Conference on Robotics and Automation}, 2018, pp. 3803--3810.

\bibitem{yu2020learning}
W.~Yu, J.~Tan, Y.~Bai, E.~Coumans, and S.~Ha, ``Learning fast adaptation with
  meta strategy optimization,'' \emph{IEEE Robotics and Automation Letters},
  vol.~5, no.~2, pp. 2950--2957, 2020.

\bibitem{yu2018policy}
W.~Yu, C.~K. Liu, and G.~Turk, ``Policy transfer with strategy optimization,''
  in \emph{International Conference on Learning Representations}, 2019.

\bibitem{yu2019sim}
W.~Yu, V.~C. Kumar, G.~Turk, and C.~K. Liu, ``Sim-to-real transfer for biped
  locomotion,'' in \emph{IEEE/RSJ International Conference on Intelligent
  Robots and Systems}, 2019, pp. 3503--3510.

\bibitem{hansen1995adaptation}
N.~Hansen, A.~Ostermeier, and A.~Gawelczyk, ``On the adaptation of arbitrary
  normal mutation distributions in evolution strategies: The generating set
  adaptation,'' in \emph{International Conference on Genetic Algorithms}, 1995,
  pp. 57--64.

\bibitem{kumar2021rma}
A.~Kumar, Z.~Fu, D.~Pathak, and J.~Malik, ``{RMA}: Rapid motor adaptation for
  legged robots,'' in \emph{Robotics: Science and Systems}, 2021.

\bibitem{yin2023risk}
J.~Yin, Z.~Zhang, and P.~Tsiotras, ``Risk-aware model predictive path integral
  control using conditional value-at-risk,'' in \emph{IEEE International
  Conference on Robotics and Automation}, 2023, pp. 7937--7943.

\bibitem{vaswani2017attention}
A.~Vaswani, N.~Shazeer, N.~Parmar, J.~Uszkoreit, L.~Jones, A.~N. Gomez,
  {\L}.~Kaiser, and I.~Polosukhin, ``Attention is all you need,''
  \emph{Advances in Neural Information Processing Systems}, vol.~30, 2017.

\bibitem{he2016deep}
K.~He, X.~Zhang, S.~Ren, and J.~Sun, ``Deep residual learning for image
  recognition,'' in \emph{IEEE Conference on Computer Vision and Pattern
  Recognition}, 2016, pp. 770--778.

\bibitem{ba2016layer}
J.~L. Ba, J.~R. Kiros, and G.~E. Hinton, ``Layer normalization,''
  \emph{Advances in NIPS 2016 Deep Learning Symposium, arXiv:1607.06450}, 2016.

\bibitem{wang2021rough}
S.~J. Wang, S.~Triest, W.~Wang, S.~Scherer, and A.~Johnson, ``Rough terrain
  navigation using divergence constrained model-based reinforcement learning,''
  in \emph{Conference on Robot Learning}, 2021.

\bibitem{hochreiter1997long}
S.~Hochreiter and J.~Schmidhuber, ``Long short-term memory,'' \emph{Neural
  Computation}, vol.~9, no.~8, pp. 1735--1780, 1997.

\bibitem{gandhi2021robust}
M.~S. Gandhi, B.~Vlahov, J.~Gibson, G.~Williams, and E.~A. Theodorou, ``Robust
  model predictive path integral control: Analysis and performance
  guarantees,'' \emph{IEEE Robotics and Automation Letters}, vol.~6, no.~2, pp.
  1423--1430, 2021.

\bibitem{chua2018deep}
K.~Chua, R.~Calandra, R.~McAllister, and S.~Levine, ``Deep reinforcement
  learning in a handful of trials using probabilistic dynamics models,''
  \emph{Advances in Neural Information Processing Systems}, vol.~31, 2018.

\bibitem{williams2018information}
G.~Williams, P.~Drews, B.~Goldfain, J.~M. Rehg, and E.~A. Theodorou,
  ``Information-theoretic model predictive control: Theory and applications to
  autonomous driving,'' \emph{IEEE Transactions on Robotics}, vol.~34, no.~6,
  pp. 1603--1622, 2018.

\bibitem{feller2016relaxed}
C.~Feller and C.~Ebenbauer, ``Relaxed logarithmic barrier function based model
  predictive control of linear systems,'' \emph{IEEE Transactions on Automatic
  Control}, vol.~62, no.~3, pp. 1223--1238, 2016.

\bibitem{kingma2014adam}
D.~P. Kingma and J.~Ba, ``Adam: A method for stochastic optimization,'' in
  \emph{International Conference for Learning Representations}, 2015.

\end{thebibliography}
\end{document}